\documentclass{article} 
\usepackage{iclr2025_conference,times}
\usepackage{booktabs}
\usepackage[shortlabels]{enumitem}
\usepackage{longtable}
\usepackage{multirow}
\usepackage{graphicx}
\usepackage{anyfontsize}
\usepackage{times}
\usepackage{latexsym}
\usepackage{booktabs}

\usepackage{amsmath,amsfonts,bm}

\def\eqref#1{equation~\ref{#1}}

\def\1{\bm{1}}

\DeclareMathAlphabet{\mathsfit}{\encodingdefault}{\sfdefault}{m}{sl}
\SetMathAlphabet{\mathsfit}{bold}{\encodingdefault}{\sfdefault}{bx}{n}

\usepackage{hyperref}
\usepackage{url}
\usepackage[export]{adjustbox}

    \title{Mitigating Copy Bias in In-Context Learning through Neuron Pruning}

\author{\thanks{Work done while visiting at The University of Edinburgh.}~~\thanks{Corresponding author} ~Ameen Ali$^{1}$  ~~
\textbf{ Lior Wolf$^{1}$
\; Ivan Titov}$^{2,3}$  \\
$^1$Tel Aviv University, 
\;$^2$University of Edinburgh,
\; $^3$University of Amsterdam
\\
\texttt{ameenali023@gmail.com, wolf@cs.tau.ac.il, ititov@inf.ed.ac.uk} \\
}

\iclrfinalcopy 
\begin{document}

\maketitle

\begin{abstract}
Large language models (LLMs) have demonstrated impressive few-shot in-context learning (ICL) abilities. Still, we show that they are sometimes prone to a `copying bias', where they copy answers from provided examples instead of learning the underlying patterns. In this work, we propose a novel and simple method to mitigate such copying bias.  First, we create a synthetic task and use the Integrated Gradients method to identify neurons that prioritize copying over generalization. We demonstrate that pruning these neurons consistently improves performance across a diverse set of ICL tasks. We also show that our method is applicable across various LLM architectures, including Transformers and State-Space Models, without requiring modifications.  
In our analysis, we adopt a task-recognition perspective on ICL and examine task vectors ~\citep{hendel2023context} induced by the model. We find that pruning enhances the quality of these vectors, suggesting that the pruned neurons previously hindered effective task recognition.

\end{abstract}
\section{Introduction}

In-Context Learning (ICL) \citep{brown2020language} has recently emerged as a powerful and simple alternative to traditional training and fine-tuning. ICL involves presenting a Large Language Models (LLM) with a ``context'' consisting of several example pairs, each containing an input and its corresponding correct output, followed by a test example for prediction. For instance, consider the following prompt:

\begin{center}
    New York $\rightarrow$ 2,~~ Lyon $\rightarrow$ 1,~~ Florida $\rightarrow$
\end{center}

In this case, the model must leverage the contextual information from the given examples to identify the underlying pattern of mapping words to vowel counts; based on this, it must then predict that the corect answer for ``Florida'' is ``3''. 

While ICL has shown considerable effectiveness, its use in few-shot scenarios faces significant challenges~\citep{zhao2021calibrate,razeghi2022impact}. In these settings, the inherent scarcity of labeled examples becomes a critical bottleneck, as ICL often requires a substantial number of in-context examples to generalize effectively. Moreover, the performance of ICL is highly sensitive to various aspects of the prompt and the presentation of the examples. Factors such as the specific wording of the prompt~\citep{wang-etal-2024-large-language-models-fair}, the order in which the examples are presented~\citep{lu2021fantastically}, and their relevance to the target example can significantly influence the outcome. Consequently, in domains where labeled data is limited, these challenges collectively hinder the reliable application of ICL, emphasizing the need for strategies that can mitigate sensitivity and make the most of the scarce examples available.

Recent research has primarily addressed these challenges by focusing on prompt formulation strategies, including techniques for selecting optimal templates and examples~\citep{zhou2023large,hao2022structured,lu-etal-2022-fantastically}, as well as calibration methods~\citep{han2023prototypical,zhao2021calibrate}. However, existing work has not yet explored how errors in in-context learning (ICL) relate to the internal processes of LLMs or how to correct them through targeted model modifications.

Our study takes a novel approach by investigating neural activation patterns related to a common challenge in ICL: copying errors. Referring back to the vowel-counting example, a copying error would occur if the model were to output ``2'' or ``1'' for Florida, instead of the correct answer ``3''.  In these cases, the model appears to directly copy an answer from the provided examples rather than generating the correct, novel response based on the induced underlying pattern.

In this work, we hypothesize that there is a small subset of neurons in language models that prioritize copying behavior over task recognition. We posit that these mechanisms can be task-agnostic; that is, the same neurons are responsible for this reasoning shortcut across a range of tasks. We further hypothesize that deactivating these neurons will make the model less likely to follow shortcuts and encourage it to focus on recognizing underlying regularities.

To identify these neurons, we employ the vowel-counting task and apply the attribution method, Integrated Gradients (IG)~\citep{sundararajan2017axiomatic}, to trace the copying errors to individual neurons. We then select the top contributing neurons as "Copying Neurons." The vowel-counting task is particularly interesting, as it appears challenging for a range of models and elicits the copying behavior in these models. We demonstrate that deactivating the neurons identified using this single task improves results across a diverse range of ICL problems, making our method practical -- since the neurons do not need to be selected for each individual task -- and confirming the existence of a general mechanism prioritizing the shortcut over reasoning.

In summary, our contributions are fourfold:
(1) We identify the copying bias in ICL and demonstrate that LLMs, particularly smaller ones, suffer from a high percentage of these errors.
(2) We introduce a method to identify specific neurons responsible for triggering the copying behavior (copying neurons).
(3) We show that pruning these identified neurons leads to out-of-the-box improvement across a wide range of ICL tasks.
(4) We utilize the task vectors framework introduced by ~\citep{hendel2023context}, quantifying a model's ability to recognize and adapt to tasks during ICL. Using this framework, we provide evidence that pruning the copying neurons enhances the quality of task vectors, indicating improved task representation. This finding explains the observed performance gains across various ICL tasks.

\section{Related Work}

ICL, first introduced by~\citep{brown2020language}, has attracted significant interest in recent years due to its ability to enable large language models to perform complex downstream tasks without explicit fine-tuning. By leveraging contextual information provided within the input prompts, these models can dynamically adapt their behavior and generate contextually relevant outputs across a wide range of tasks. While it is commonly associated with Transformer architectures, ICL has also been explored in other model architectures, such as State-Space Models and the RWKV model~\citep{grazzi2024is,park2024can}. Leading to a wide line of works that seek to improve the effectiveness of ICL mechanisms~\citep{zhao2021calibrate,weisymbol,chu2023fine,zhou2023leasttomost,cot,li2024mend}, as well as studies that aim to explain the underlying processes and dynamics of how models internalize and utilize context~\citep{min2022rethinkingroledemonstrationsmakes,liu-etal-2022-makes,xie2022an,olsson2022context,von2023transformers,dai2022can}.

Works proposing methods to improve ICL have mainly focused on prompt selection and prompt formulation strategies,~\citet{zhou2023leasttomost} propose a different prompting strategy that breaks down a complex problem into a series of simpler subproblems and then solves them in sequence. \citet{zhang-etal-2022-active} reformulate the example selection for ICL as a sequential decision problem, and propose a reinforcement learning algorithm for identifying generalizable policies to select demonstration examples. \citet{sorensen2022information} introduces a method for unsupervised selection of prompt templates by maximizing the mutual information between the input and the corresponding model output. \citet{lu-etal-2022-fantastically} show that the order in which the samples are provided can make a significant difference and propose a method for finding the optimal permutation. 

While the majority of approaches to improving ICL have focused on various methods of prompt engineering, such as prompt selection and formulation strategies, our work takes a fundamentally different approach. To the best of our knowledge, this is the first attempt to enhance ICL through a neuron-level analysis, specifically by identifying and pruning neurons that contribute to copying (which we define in Section~\ref{sec:mem_errs}) within large language models.

Neuron-Level analysis involves examining individual neurons within the model to determine their specific roles. Relevant studies in this area have aimed to understand and categorize neurons into groups based on their functional roles. For example, \citet{voita2023neurons} show that individual neurons in LLMs correspond to different groups such as dead neurons, positional neurons, and N-gram neurons. Furthermore, ~\citet{gurnee2024universal} discusses an additional group of categories including entropy neurons, alphabet neurons, syntax neurons, and semantic neurons. \citet{chen2024finding} identify safety neurons, which are responsible for safety behaviors. Neuron-level analysis was further expanded to study multilingual LLMs. ~\citet{tang-etal-2024-language} detect language-specific and language-agnostic related neurons in multilingual language models. Neuron-Level analysis is typically performed through activation analysis, where one examines the patterns of neuron activations across various inputs~\citep{voita2023neurons,gurnee2024universal,stolfo2024confidence}. Attribution methods, such as Integrated Gradients~\citep{sundararajan2017axiomatic}, are also employed to quantify the contribution of individual neurons to the model's output, hence, allowing for discovering different neuron-families~\citep{dai-etal-2022-knowledge,shi2024ircan}. 

\section{Method}
This section presents our method for detecting and mitigating copying in ICL. First, in section~\ref{sec:prem} we revisit the ICL setting, formally define what are the copying errors, and introduce the Integrated Gradients attribution method. In section~\ref{sec:syntheic} we elaborate on how we create a synthetic dataset that will be used as a proxy for our proposed detection method, and in section~\ref{sec:method} we present our method for detecting and mitigating copying neurons.
\subsection{Preliminaries}
\label{sec:prem}
\paragraph{In-Context Learning}
ICL enables a model $f$ to adapt to downstream tasks without any parameter updates. This is achieved by forming a prompt $p$ that includes concatenated training examples. In ICL, a prompt $p$ is constructed by linking the task inputs with their corresponding labels as follows: 
\begin{equation}
    p = x_1 : y_1,  x_2 : y_2, \dots x_n : y_n , x_{n+1} : 
\end{equation}
 Using this prompt, the model $f$ predicts the most probable label $y$ that completes the prefix $p$ according to the function $f$. In this framework, few-shot learning is characterized by the number of examples in the prompt. Furthermore, we denote $S = [y_1, y_2, \dots y_n]$ as the set of the in-context example answers for prompt $p$, then we say  $p$ is an $|S| = n$-shot in-context prompt.

\paragraph{Copying Bias}
 \label{sec:mem_errs}
Copying bias, as we define it, refers to the phenomenon where a language model $f$ returns an incorrect answer that is one of the labels $S$ of the in-context samples provided in the prompt. In other words, given a prompt $p$ under the $n$-shot ICL settings, a prediction $y_{n+1}$ is called a copying prediction if (1) : $y_{n+1}$ is a wrong prediction and (2) : $y_{n+1} \in S$. 

We hypothesize that there exists a small number of neurons, which we call copying neurons, that trigger the model to copy responses from the prompt examples $S$ of the prompt $p$. The identification of these neurons is, therefore, crucial for understanding how LLMs balance copying and generalization, and for enhancing the reliability and interpretability of these models. We further hypothesize that pruning these neurons by setting their weights to zero would encourage the model to reason rather than solely rely on copying, thereby improving its ability to generalize under few-shot ICL.

\paragraph{Integrated Gradients (IG)}

Integrated Gradients (IG) is a popular technique in explainable AI~\citep{sundararajan2017axiomatic} used to elucidate the relationship between a model's predictions and its input features. It applies to any differentiable model and is computationally efficient. IG works by generating attributions based on integrating the gradients as the input varies between a \textit{baseline} and the final input of interest (the path):
\begin{equation} 
    \textbf{IG}(f, \tilde{x}, x, i) =
    (x_i - \tilde{x}_i) \int_{\alpha = 0}^1\frac{\partial f(\hat{x})\cdot d\alpha}{\partial f(\hat{x_i})}\Bigg|_{\hat{x} = \tilde{x} + \alpha (x - \tilde{x})}\,,
\end{equation}

where $f({\cdot})$ denotes the prediction of the model, $x$ is the input vector of interest that we want to attribute, $\tilde{x}$ is the baseline input vector and $i$ is an index denoting the indices of features of interest. The baseline represents a reference point against which the input of interest is compared. More specifically, the baseline is an input vector that is supposed to reflect a neutral, missing, or reference state. The idea behind using a baseline is to measure how the model's output changes as we transition from this baseline state to the actual input of interest. This change is quantified by integrating the gradients along this path. In our experiments, we use the constant zero baseline as proposed in the original paper of IG. This baseline is straightforward to implement and often provides meaningful attributions. The integral is practically computed using the Riemann sum approximation:
\begin{equation}
\textbf{IG}(f, \tilde{x}, x, i) \approx \frac{(x_i - \tilde{x}_i)}{m} \sum_{r=1}^{m} \frac{\partial f(\tilde{x} + \frac{r}{m}(x - \tilde{x}))}{\partial x_i},
\end{equation}
where $m$ is the total number of Riemann steps.

\subsection{Proxy ICL Dataset Generation}
\label{sec:syntheic}
The Integrated Gradients (IG) method operates by backpropagating the probability of the prediction. To effectively utilize IG within our framework, we require a set of in-context examples with ground-truth labels. We avoid relying on access to target task data for neuron selection, as this would limit the practicality of our approach. Instead, we utilize a synthetic proxy dataset to identify neurons and prune the same set of neurons on evaluation tasks. This approach also aligns with our research hypothesis that copying neurons are largely task-agnostic. This synthetic dataset is employed within the Integrated Gradients framework to identify copying neurons. 

The specific task we choose is that of vowel counting since the mapping from a word to such structural attributes requires reasoning. LLMs can occasionally make errors on this task and similar tasks,\footnote{See, e.g., \tiny{\url{https://community.openai.com/t/incorrect-count-of-r-characters-in-the-word-strawberry/829618}}.} potentially outputting copying responses. The synthetic samples we utilize simply map an arbitrary word to its corresponding vowel counts (e.g., \textit{apple}: $2$). To construct a diverse set of examples, we extract words from a dictionary and calculate their respective vowel counts. This results in two sets: a training set used to identify the copying neurons and a validation set employed to select the optimal layer and percentage of neurons for intervention through pruning.

\subsection{Copying Neurons Detection}
\label{sec:method}
 \begin{figure*}[t]
\centering
\includegraphics[width=0.8\textwidth,height=0.5\textwidth]{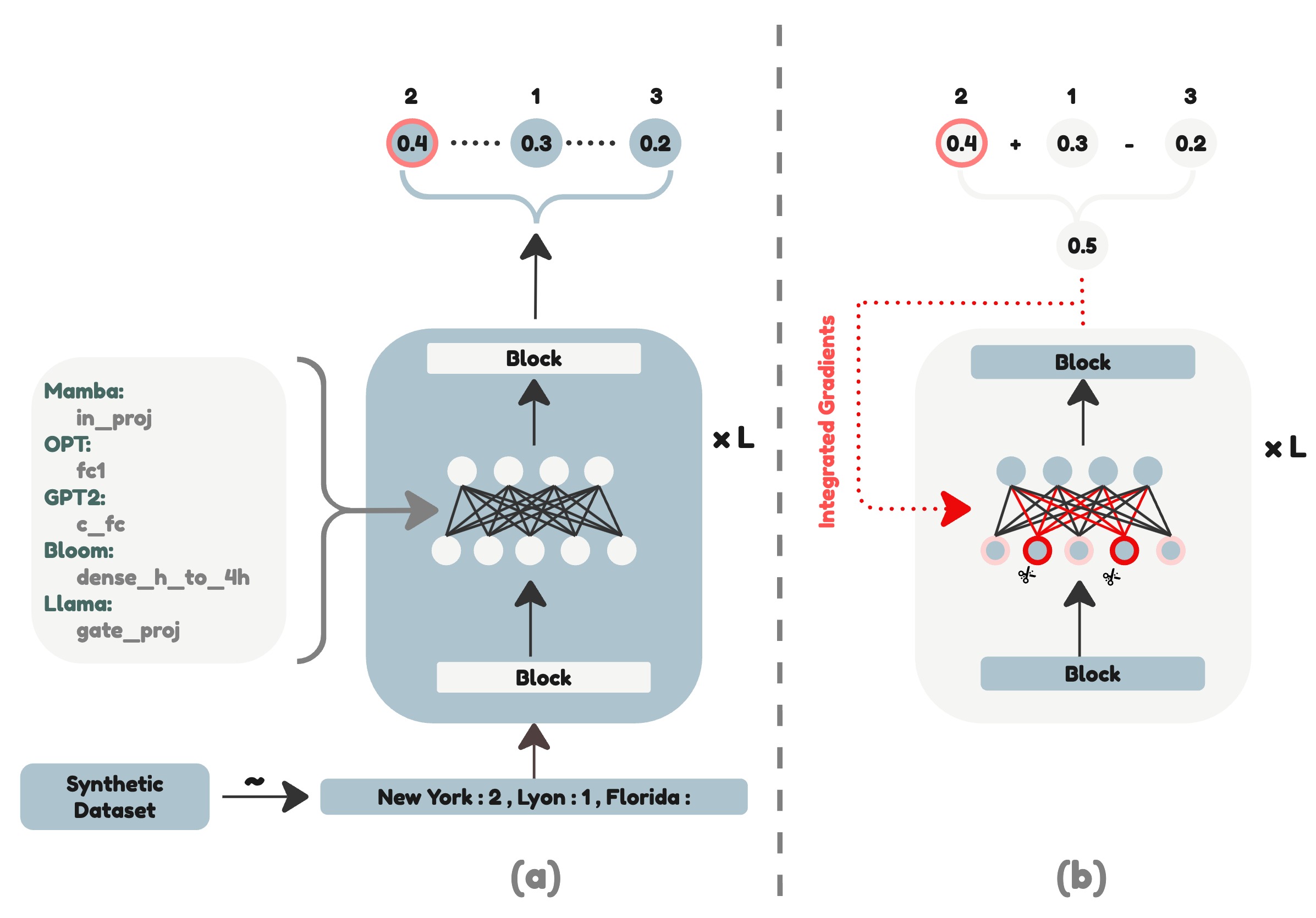}
\caption{A high-level depiction of our proposed method of detecting copying neurons. First, in (a) we feed ICL prompts from the synthetic dataset. In this phase, we are only interested in the prompts where the model outputs a wrong response which also appears in the prompt examples. Second, in (b) we use these prompts and calculate the sum of the probabilities over predicted responses that appear in the prompt. This sum is used within the IG framework to attribute it to neurons in the targeted layer.}
\label{fig:viewsOfLLMs}
\end{figure*}
\label{sec:detection}
Denote $V$ as the vocabulary space, $p$ as the in-context prompt of interest containing $n$ in-context examples, and $S_p = [y_1, \dots y_n]$ the set of labels of the different examples in the prompt $p$. In our detection process, we are only interested in prompts $p$ on which the model outputs a wrong prediction $y \in S_p$ (hence the copying) and denote $\hat{y} \notin S_p$ as the ground-truth answer. Let $w^l \in \mathrm{R}^{d_1\times d_2}$ be the weight matrix of the linear layer at block $l$ on which our detection process operates. Furthermore, we define the model output $\mathrm{P}_p(y | \hat{w}^{l}_{j})$ as the probability of predicting a certain answer $y  \in \mathrm{V}$.
\begin{equation}
    \mathrm{P}(y | \hat{w}^{l}_{j}, p) = \mathrm{P}(y| w^{l}_{j}=\hat{w}^{l}_{j}, p), 
\end{equation}
where $w^{(l)}_{j}$ denotes the $j$-th intermediate neuron in the $l$-th layer of interest (Figure~\ref{fig:viewsOfLLMs}), 
$\hat{w}^l_j$ is a given constant that $w^l_j$ is assigned to. We define $l_u = \sum_i^n\mathrm{P}(y = y_i \in S_p | w^l_j = \hat{w}^l_j, p)$ as the probability of predicting an answer which is provided in the prompt examples (in $S$), we also define $l_v = \mathrm{P}(y=\hat{y} \notin S_p | w^l_j = \hat{w}^l_j, p)$ as the probability of predicting the ground-truth answer $\hat{y}$. Lastly, we define $\Delta L= l_v - l_u$ as the prediction shift. 

Copying, by definition, occurs when the model's prediction shifts from the true answer to one of the responses provided in the prompt. Thus, copying neurons are those that drive $\Delta L$.

By leveraging IG, we can attribute $\Delta L$ to individual components. This approach enables us to identify specific neurons responsible for copying within the LLM.

To quantify the contribution of a neuron $w_j^l$ to the  prediction shift ($\Delta L$), we gradually change $w^{(l)}_{j}$ from $0$ (the \textit{baseline}) to its original value $\hat{w}^{(l)}_{j}$ computed by the model and integrate the gradients: 
\begin{align}
& \text{Attr}(w^{l}_{j}, p, S_p) = \mathrm{IG}(w^{l}_{j}) =  (\hat{w}^{l}_{j}-0) \int_{\alpha=0}^{1} \frac{\partial |\Delta L|}{\partial w^{l}_{j}} d \alpha  \nonumber \\ &\qquad\qquad\qquad = \sum_i^n \hat{w}^{(l)}_{j} \int_{\alpha=0}^{1} \frac{\partial \left|\mathrm{P}(y_i | \alpha \hat{w}^{l}_{j}, p) - \mathrm{P}(\hat{y} | \alpha \hat{w}^{l}_{j}, p)\right|}{\partial w^{l}_{j}} d \alpha,
\end{align}

\begin{equation}
\text{Attr}(w^{l}_{j}, p, S_p) = \mathrm{IG} (w^{l}_{j}) =
\sum_i^n\frac{\hat{w}^{l}_{j}}{m} \sum_{r=1}^{m} \frac{\partial \left|\mathrm{P}(y_i | \frac{r}{m} \hat{w}^{l}_{j}, p) - \mathrm{P}(\hat{y} | \frac{r}{m} \hat{w}^{l}_{j}, p)\right|}{\partial w^{l}_{j}},
\end{equation}
where $m=20$ is the number of approximation steps we use in our experiments, following \citep{sundararajan2017axiomatic}.

Finally, we compute the final attribution scores for the neurons $[w_j^{l}]$ by averaging $\text{Attr}(w_j^{l})$ across all samples in the synthetic dataset, resulting in a relevance score that quantifies the extent to which a neuron contributes to copying.
\begin{equation}
\label{eq:final_ig}
\text{R}(w_j^{l}) = \frac{1}{|D|}\sum_{k=1}^{|D|} \text{Normalize}(\text{Attr}(w_j^{l}, p_k, S_{p_k})),
\end{equation}
where $D$ is the synthetic dataset and $p_k$ is the $k-$th sample of the synthetic dataset. and Normalize is the min-max normalizing function.

To mitigate the copying bias, we propose to suppress the weights of the detected copying neurons as follows:
\begin{equation}
    w^l_i = 
      \left\{
  \begin{aligned}
    0, &\qquad& \text{R}(w^l_i) \geq \sigma, \\
    w^l_i, &\qquad&   \text{R}(w^l_i) < \sigma.
  \end{aligned}
  \right.
\end{equation}
where $\sigma$ is the filtering threshold and is tuned using a validation dataset.

\section{Experiments}

 \begin{figure*}[t]
\centering
\includegraphics[width=1\textwidth,height=0.7\textwidth]{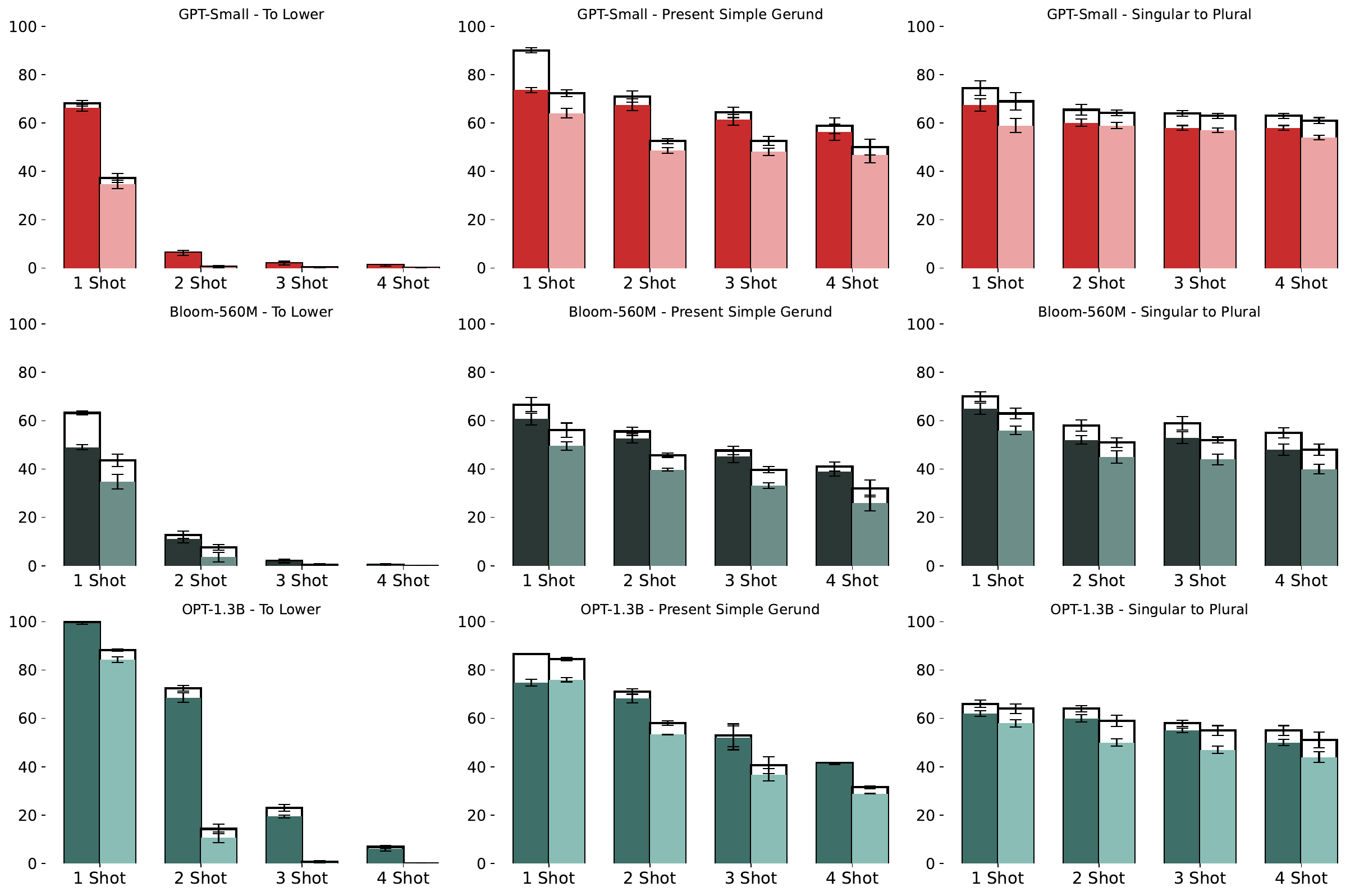}
\caption{Percentage of total errors and copying errors for both the pruned and un-pruned models, results are shown for 3 ICL tasks across 3 different models: GPT2-Small, BLoom-560M, and OPT-1.3B. The dack bar in each diagram represents the unpruned version while the lighter bar represents the pruned version; the entire bar height represents the total error of the model and the shaded part represents the copying error rate.}
\label{fig:mem_error_analysis}
\end{figure*}

Our approach offers a generic method for detecting copying neurons, applicable to any language model. To demonstrate the generalizability of our method across different models and tasks, we conduct extensive experiments on a diverse set of LLMs. This includes the recent state-space models, such as Mamba~\citep{mamba} as well as a broad spectrum of transformer-based models, including OPT~\citep{zhang2022opt}, GPT2~\citep{radford2019language}, Bloom~\citep{le2023bloom} and LLaMA~\citet{touvron2023llama,dubey2024llama}.
\paragraph{Data} In all of our experiments, we follow \citet{hendel2023context} and \cite{grazzi2024mamba} and study 18 tasks in 3 different categories including algorithmic (to lowercase, to uppercase, list first, list last, list max, list min, and next letter), linguistic (present to past, present to gerund, singular to plural, antonyms, and past to perfect), and knowledge (landmark, currency, country to capital, person to language, religion, and continent), the algorithmic tasks are generated automatically, for the linguistic we use the GitHub Repositries\footnote{\url{https://github.com/Drulac/English-Verbs-Conjugates}}\footnote{\url{https://github.com/sindresorhus/irregular-plurals}} and the knowledge data is taken from~\cite{meng2022locating}. Additionally, we also incorporate real-world datasets like sentiment classification, including SST2, SST5, and subsets from the BIG-Bench Tasks~\citep{suzgun2022challenging}. For more information about the data, refer to Appendix~\ref{appx:tasks}.
\paragraph{Implementation Details}
For each model, we use the synthetic validation dataset introduced in Section~\ref{sec:syntheic} to identify the optimal block and the number of copying neurons to prune as follows. IG, as defined in Equation \ref{eq:final_ig}, is applied across the layers of interest (summarized in Appendix.~\ref{appx:layers}) in all of the blocks in the model. As described in Section.~\ref{sec:method}, this procedure quantifies which neurons contribute most significantly to the copying errors. Furthermore, we use the validation set introduced in Section~\ref{sec:syntheic} in order to find the optimal pruning rate and the optimal block that maximizes the validation accuracy over the proxy ICL validation set, we apply multiple pruning rates ranging from 1\% to 10\%, in 1\% increments (i.e., [1\%, 2\%, 3\%, \dots, 10\%]) over each layer in all of the blocks in the model, and use the best validation accuracy performing configuration to use for the unseen ICL tasks. This procedure allows us to determine the optimal block and the optimal number of neurons to prune for maximizing the accuracy on the validation proxy ICL set. The layers we choose to apply the detection and pruning procedures are summarized in Appendix~\ref{appx:layers}.
\begin{figure}[t]
\centering
\includegraphics[width=1\textwidth,height=0.75\textwidth]{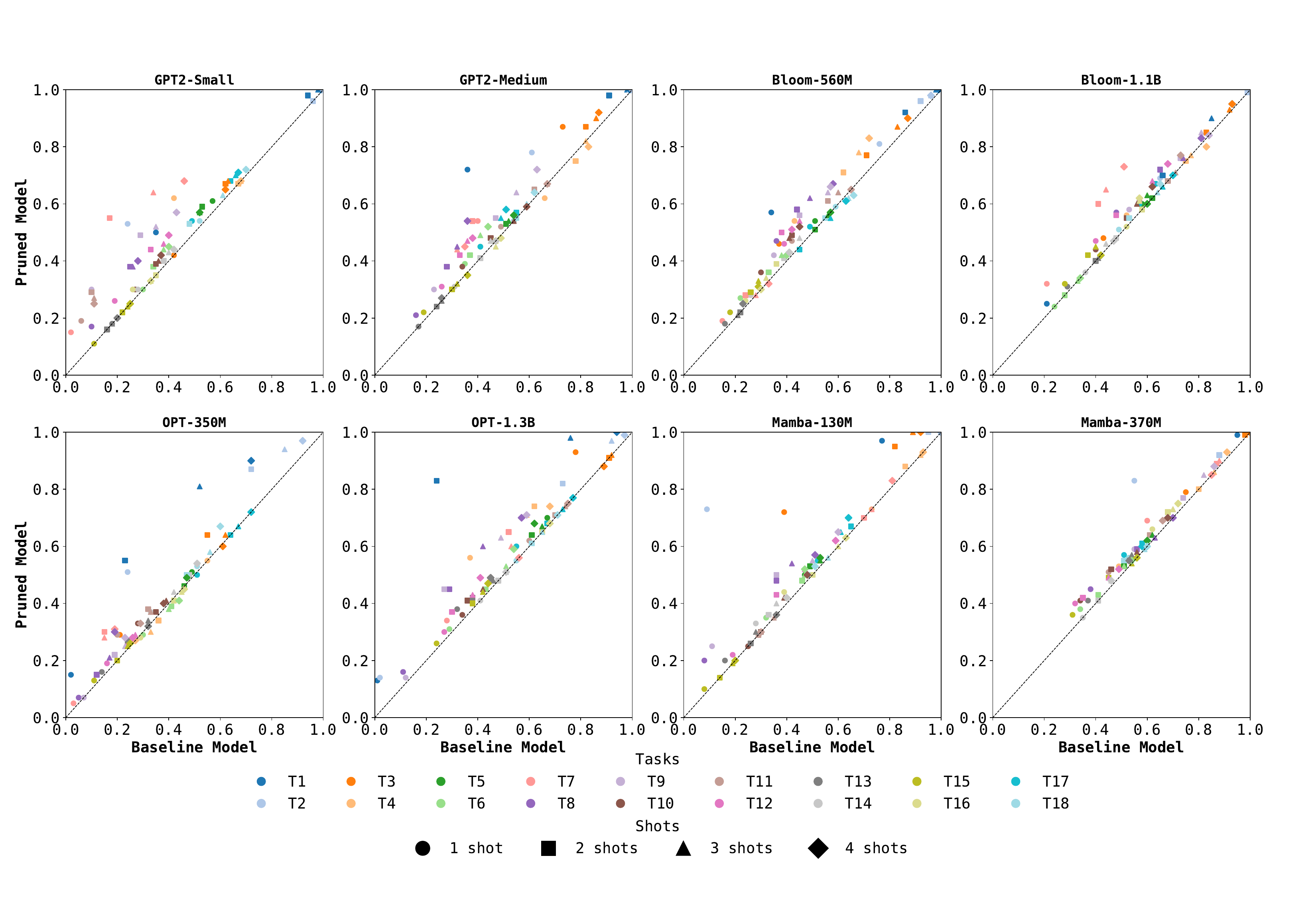}
 \vspace{-4pt}
 \caption{Summary of the results over the synthetic ICL tasks, for more information on the tasks and the exact numbers, refer to Appendix~\ref{appx:results}.}\label{fig:results}
\vspace{-7pt}
\end{figure}

\begin{figure*}[t]
\centering
\includegraphics[width=1\textwidth,height=0.27\textwidth]{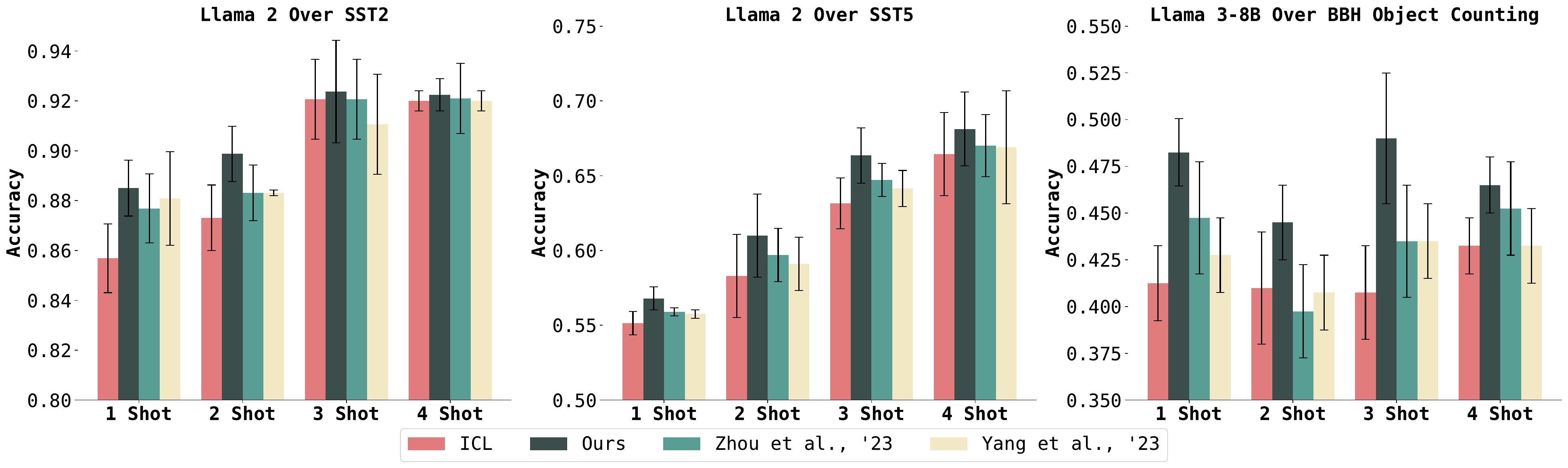}
 \caption{Results of Llama-2 and Llama-3 over SST2, SST5, and Object Counting task from BBH benhmark}\label{fig:llama}
\end{figure*}
\subsection{Validating the Prevalence of Copying Errors}

To validate the significance of copying errors, we first conduct an error analysis for three representative ICL tasks (to lower, singular to plural, and present simple to gerund) across a range of large language models (GPT2-Small, Bloom-560M, and OPT-1.3B). In this analysis, we show the percentage of copying errors as defined in Section~\ref{sec:mem_errs} out of the total number of errors. The results are presented in  Figure~\ref{fig:mem_error_analysis}. In this figure, we present a comparative analysis of error rates across different models and tasks. The dark bars show error rates for the baseline (unpruned) model, while the light bars display results for the pruned model. Each bar is composed of two elements: an unfilled portion representing the total error rate, depicted by the full height of the bar with a black outline, and a filled portion below indicating the copying error rate, which is a subset of the total error rate. Evidently, most of the errors in these few-shot ICL tasks stem from copying, these models tend to replicate responses based on examples provided in the prompt, rather than generalizing to new contexts. Additionally, our pruning method significantly reduces the number of copying errors thus also reducing the total error rate.

\subsection{Tasks Evaluation}
To demonstrate the generalizability of our approach across model architectures, we include a range of models: (1) Transformer-based: OPT, GPT-2, BLOOM, LLaMA, and (2) Mamba state space models of various sizes.

For each synthetic ICL task, we utilize the test sets introduced by~\citet{hendel2023context}. We report the mean accuracy across these test sets for each model and task configuration over the different shots (1, 2, 3, and 4) evaluated using various seeds. Figure.~\ref{fig:results} presents a scatter plot comparing the performance of our pruned model against the baseline (non-pruned) model. A diagonal line representing equal performance is included for reference. Data points falling above this line indicate instances where our pruned model achieves higher accuracy than the baseline, while points below the line represent cases where the baseline model outperforms. As evident from the distribution of points that is dominated by those above the diagonal, our pruned model consistently demonstrates superior accuracy across a wide range of ICL tasks for different shot instances, underscoring the effectiveness of our targeting neuron pruning strategy. We believe that the fact our technique rarely leads to a performance drop -- and when it does, the impact is only marginal -- makes it particularly appealing for practical applications. For the exact numbers, we refer the reader to Appendix~\ref{appx:results}.

In order to test our approach beyond this benchmark ICL tasks, we also test it on tasks that are based on three datasets of collected data: SST-2, SST-5, and the object counting sub-task from the BBH benchmark~\citep{suzgun2022challenging}. In these cases, to allow a comparison with previous work, we use Llama-2~\citep{touvron2023llama} and Llama-3~\citep{dubey2024llama}.  The same synthetic dataset of vowels mapping is used for copying neurons detection. 

We include two recent baselines. The first is Weighted In-Context Learning (WICL) by ~\citet{yang-etal-2023-demonstration}, which improves the performance of LLMs on downstream tasks by assigning and applying optimal weights to demonstration examples during ICL. The second, Automatic Prompt Engineer (APE) by~\citet{zhou2023large}, automatically generates and selects optimal instructions for large language models to improve their performance on various ICL tasks without relying on human-crafted prompts. 

The results for these three benchmarks are presented in Figure~\ref{fig:llama}. Evidently, our method significantly improves over the baseline LLM as well as over the two baselines. 

\subsection{Tasks-Vector Analysis}
Next, we build upon the recent task-vectors framework of~\citet{hendel2023context} to study the relationship between copying neurons pruning and the quality of the emerged task vectors in ICL. By comparing the task vectors generated by pruned and unpruned models across various ICL scenarios, we seek to understand if our proposed targeted pruning can enhance a model's ability to distill task-relevant information from demonstrations, specifically under the few shots settings. Furthermore, we follow the setup of~\citet{hendel2023context} and report ICL accuracy using the standard ICL promoting (denoted by ICL), the accuracy obtained by using the emerged Task-Vectors without pruning the copying neurons (denoted as Task-Vectors) and the accuracy obtained by using Task Vectors obtained from the model with the pruned copying neurons (denoted as Tak-Vectors-Pruned). 

\begin{figure*}[h]
\centering
\includegraphics[width=1\textwidth,height=0.35\textwidth]{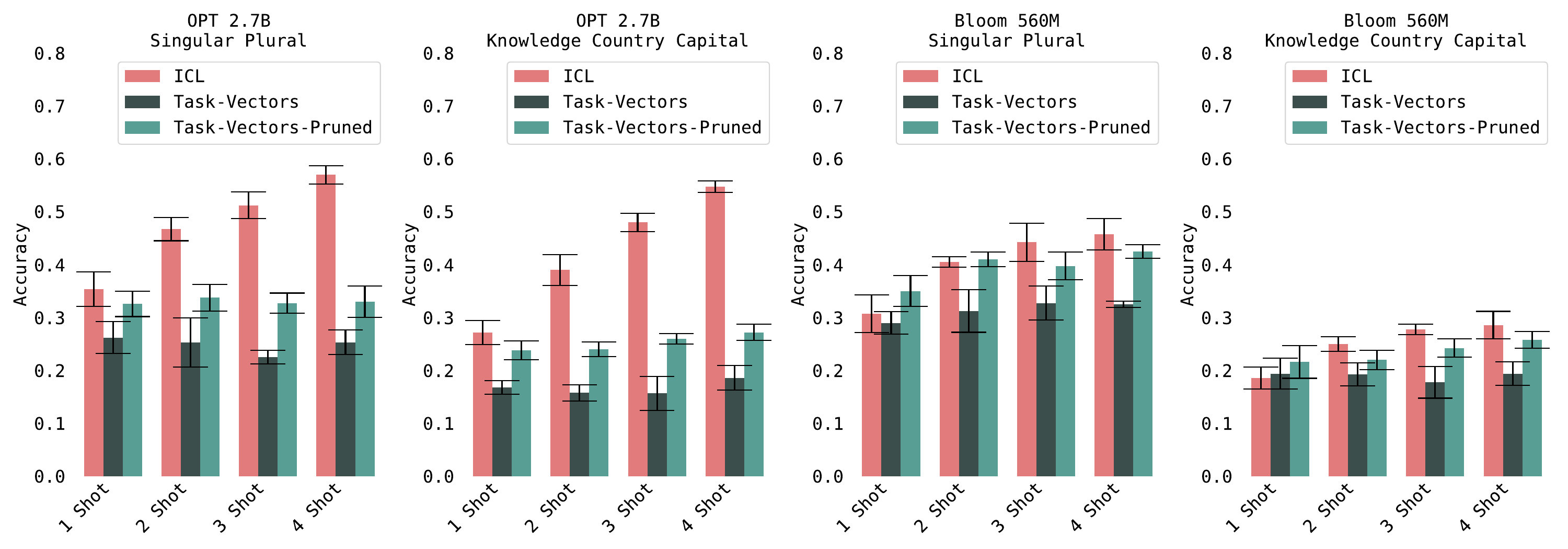}
\caption{Task-Vectors accuracies over OPT-2.7B and Bloom-560M models tested on (1) Singular Plural and (2) Country Capital ICL tasks. We show the Task-Vectors accuracies with and without pruning the detected copying errors, as can be seen, pruning the copying errors improves the quality of the extracted Task-Vectors across the different shots $\in [1, 2, 3, 4]$ for the two models and ICL tasks.}
\label{fig:task_vectors}
\end{figure*}

In Figure.~\ref{fig:task_vectors}, we show the results of OPT-2.7B and Bloom-560M models over the  ``Singular Plural'' and ``Country Capital'' ICL tasks. As can be seen from the results, pruning the copying neurons indeed yields better Task-Vectors for ICL. This suggests that our pruning strategy may be effectively identifying and removing neurons that were interfering with the model's ability to infer the underlying task. For more results on additional models and tasks, refer to Appendix~\ref{appendix:taskvectors}
.

\subsection{Ablation Studies}

We present multiple ablation studies to evaluate and understand the different components of our proposed detection and pruning methods. These ablation studies were conducted with the OPT-350M, GPT2-Small, and Bloom-560M models, over the Linguistic Antonyms, and Letter to Upper ICL tasks.

Our first experiment focuses on using ``Prediction Shift'' within the IG framework. We aim to determine whether applying IG to the prediction shift, as defined in Section~\ref{sec:method}, is essential for our proposed method's effectiveness. To this end, we compare our approach with an alternative where IG is applied to the predicted probability (specifically, the maximum probability) instead. Results are presented in Tab.\ref{table:ablations} (``Max IG'' row) clearly shows that the prediction shift is essential for the success of our proposed method. 

We further check the effect of min-max normalizing the IG scores across the samples from the proxy ICL task we use in the detection process. The results without this normalization are reported under ``w/o Norm''. As can be seen, normalizing the scores across the samples can significantly enhance the detection process of the copying neurons.

Additionally, we explore a baseline case where we randomly prune the same percentage of neurons as in the best-performing version of our method. This experiment, labeled as ``Random'' in Tab.\ref{table:ablations}, shows a degradation in ICL accuracy for some shot settings, underscoring the importance of our targeted pruning strategy. 

\begin{table}[t]

\centering
\scriptsize
\setlength{\tabcolsep}{6pt}
\begin{tabular}{l|ccccc|ccccc|ccccc}
\toprule
& \multicolumn{5}{c|}{\textbf{OPT-350M}} & \multicolumn{5}{c|}{\textbf{GPT-Small}} & \multicolumn{5}{c}{\textbf{Bloom-560M}} \\
\rotatebox{90}{\textbf{Shot}} & 
\rotatebox{90}{ICL} & 
\rotatebox{90}{Ours} & 
\rotatebox{90}{Max IG} & 
\rotatebox{90}{w/o Norm} & 
\rotatebox{90}{Random} & 
\rotatebox{90}{ICL} & 
\rotatebox{90}{Ours} & 
\rotatebox{90}{Max IG} & 
\rotatebox{90}{w/o Norm} & 
\rotatebox{90}{Random} & 
\rotatebox{90}{ICL} & 
\rotatebox{90}{Ours} & 
\rotatebox{90}{Max IG} & 
\rotatebox{90}{w/o Norm} & 
\rotatebox{90}{Random} \\
\midrule
\multicolumn{16}{c}{\textbf{Linguistic Antonyms}} \\
\midrule
1 & 0.20 &\textbf{0.29} & 0.18 & 0.22 & 0.20   & 0.06 & \textbf{0.19} & 0.17 & 0.15 & 0.08 & 0.57 & \textbf{0.61} & 0.55 & 0.59 &  0.57\\
2 & 0.32 & \textbf{0.38} & 0.30 & 0.34 & 0.30   & 0.10 & \textbf{0.29} & 0.29 & 0.21 & 0.12 & 0.68 & 0.68 & 0.68 & 0.68&  0.67\\
3 & 0.33 & \textbf{0.37} & 0.33 & 0.34 & 0.30   & 0.11 & \textbf{0.27} & 0.25 & 0.23 &  0.11& 0.71 & 0.71 & 0.70 &  0.70& 0.70 \\
4 & 0.29 & \textbf{0.33} & 0.27 & 0.31 & 0.27   & 0.11 & \textbf{0.25} &  0.25& 0.22 & 0.13 & 0.73 & \textbf{0.77} & 0.73 & 0.77 & 0.75 \\
\midrule
\multicolumn{16}{c}{\textbf{Letter to Uppercase}} \\
\midrule
1 & 0.24 & \textbf{0.51} & 0.30 & 0.37 & 0.23 & 0.24 & \textbf{0.53} & 0.37 & 0.45 & 0.34 & 0.76 & \textbf{0.81} & 0.77 & 0.78 & 0.76 \\
2 & 0.72 & \textbf{0.87} & 0.72 & 0.81 & 0.72 & 0.96 & 0.96 & 0.95 & 0.96 & 0.96 & 0.92 & \textbf{0.96} &0.91 & 0.95 & 0.90 \\
3 & 0.85 & \textbf{0.94} & 0.85 & 0.90 & 0.84 & 1.00 & 1.00 & 0.98 & 0.99 & 1.00 & 0.96 & \textbf{0.98} & 0.96& 0.96 &  0.96\\
4 & 0.92 & \textbf{0.97} & 0.91 & 0.94 & 0.92 & 1.00 & 1.00 & 0.99 & 1.00 & 1.00 & 0.96 & \textbf{0.98} & 0.95 & 0.95 &  0.95\\
\bottomrule
\end{tabular}
\caption{Ablation studies for the different components in our method over OPT-350M, GPT2-Small, and Bloom-560M models applied on the Linguistic Antonyms, and Letter to Upper ICL tasks}

\label{table:ablations}
\end{table}

\section{Conclusions and Discussion}

We presented a novel method to mitigate copying bias in few-shot In-Context Learning by pruning neurons that are linked to this behavior according to the integrated gradients interoperability method. Our approach consistently improved performance across a variety of ICL tasks and model architectures. These findings highlight the potential of targeted neuron pruning as an effective strategy for optimizing the capabilities of large language models.

The ``out-of-the-box'' improvements provided by our method, without the need for task-specific data or fine-tuning, have significant practical implications for deploying more reliable few-shot learning systems. Our approach allows for the enhancement of LLM performance across a wide range of tasks using only a simple, synthetic dataset for neuron identification. Moreover, the consistent improvements observed across different model architectures suggest that this method could be broadly applicable, potentially becoming a standard post-processing step in LLM deployment pipelines.

The success of our pruning method in improving performance across various tasks indicates that ``copying neurons'' may be acting as a form of shortcut, inhibiting the model's ability to engage in more sophisticated reasoning processes. This observation aligns with recent work on shortcut learning in neural networks~\citep{yom-din-etal-2024-jump,belrose2023eliciting} and suggests that in-context learning quality could potentially be improved by carefully modulating the influence of different neuron groups or pathways. 

Our results suggest that by pruning copying neurons, we enhance the model's ability to distill task-relevant information from demonstrations, leading to more effective task vectors. This raises interesting questions about the relationship between neuron-level representations and the higher-level task embeddings captured by task vectors. Specifically, it may be useful to consider the representation in a way that disentangles multiple activation pathways.

\subsection*{Acknowledgements}

Ivan Titov is supported by the Dutch National Science Foundation (NWO Vici VI.C.212.053). This work was supported by a grant from the Tel Aviv University Center for AI and Data Science (TAD). This research was also supported by the Ministry of Innovation, Science \& Technology, Israel (1001576154) and the Michael J. Fox Foundation (MJFF-022407). The contribution of the first author is part of a  PhD thesis research conducted at Tel Aviv University.

\bibliography{arxiv}
\bibliographystyle{arxiv}

\appendix
\newpage
\section{Target Layers} 
\label{appx:layers}
This section outlines the exact layers targeted by our detection and pruning techniques. We concentrate on specific linear layers within each model block, with GPT2 being an exception where we focus on a CNN layer. Our approach encompasses both transformer-based and Mamba-based architectural designs. For a detailed breakdown of the exact layers our method operates on across various model families, refer to Table~\ref{table:layers}.

\begin{table*}[h]
\label{table:layers}
\caption{A summary of the specific layers on which we apply our detection and pruning method for different model families}
\footnotesize
\centering
\begin{tabular}{lll}
\textbf{Family} &\textbf{Layer} &\textbf{type} \\
\midrule
OPT & fc1 & linear\\
GPT2 & mlp.c\_fc & cnn\\
Bloom & mlp.dense\_h\_to\_4h & linear\\
Llama & mlp.gate\_proj & linear\\
Mamba & mixer.in\_proj & linear\\
\bottomrule
\end{tabular}
\end{table*}
\section{Tasks Datasets} 
\label{appx:tasks}
In all of our experiments, we follow \citet{hendel2023context} and \citet{grazzi2024mamba} and study 18 tasks in 4 different categories including algorithmic, translation, linguistic, and knowledge, the algorithmic tasks are generated automatically, for the linguistic we use the GitHub Repositries\footnote{\url{https://github.com/Drulac/English-Verbs-Conjugates}}\footnote{\url{https://github.com/sindresorhus/irregular-plurals}} and the knowledge data is taken from~\cite{meng2022locating}. More details on the datasets are shown in Table~\ref{tab:tasks}.

\begin{table*}[h]
\label{tab:tasks}
\footnotesize
\centering
\begin{tabular}{llll}
\textbf{Category} &\textbf{Reference}& \textbf{Task}  & \textbf{Example} \\
\midrule
\multirow{7}{*}{Algorithmic}
 &T1& To Lowercase &  A $\rightarrow$ a \\
 &T2& To Uppercase &  a $\rightarrow$ A \\
 &T3& List First &  q,b,e,s $\rightarrow$ q\\
 &T4& List Last &  q,b,e,s $\rightarrow$ s \\
 &T5& List Max &  2,1,5 $\rightarrow$ 5 \\
 &T6& List Min &  2,1,5 $\rightarrow$ 1 \\
 &T7& Next Letter &  a,b,c $\rightarrow$ d \\

\midrule
\multirow{5}{*}{Linguistic}
 &T8& Present to past &  go $\rightarrow$ went \\
 &T9& Present to gerund &  go $\rightarrow$ going \\
 &T10& Singular to plural &  cat $\rightarrow$ cats \\
 &T11& Antonyms  &  happy $\rightarrow$ sad \\
 &T12& Past to Perfect  &  catch $\rightarrow$ caught\\
\midrule
\multirow{5}{*}{Knowledge}
 &T13& Landmark & Maybach $\rightarrow$ Germany \\
 &T14& Currency & Azerbaijan $\rightarrow$  Manat\\
 &T15& Country to Capital &  France $\rightarrow$ Paris \\
 &T16& Person to Language &  Macron $\rightarrow$ French \\
 &T17& Religion &  Muhammad $\rightarrow$ Islam \\
 &T18& Continent & Swanson Mountains $\rightarrow$ Antarctica\\

\bottomrule
\end{tabular}
\end{table*}
Beyond synthetic ICL tasks, we use sentiment classification datasets like SST2, and SST5, in SST2 the task is to classify text sentences into one of the two sentiments (negative or positive), while in SST5 the task is to classify text sentences into one of five sentiments (very positive, positive, neutral, negative, very negative). Additionally, we also incorporate the object-counting task from the BBH benchmark~\citep{suzgun2022challenging}, where the task is to find out the total number of objects given in a context sentence, a sample illustration from the dataset is as follows:
\\

\centerline{``I have a car, and a toaster. How many objects do I have? $\rightarrow$ 2''}

\newpage
\section{More Results}
This section presents the full results of the ICL tasks. We compare the performance of various models across 18 ICL tasks. Each entry in the result tables contains two values: the left value represents the performance of the unpruned model, while the right value shows the performance achieved using our proposed method. Results are averaged across 5 different seeds.
\label{appx:results}
\begin{table*}[h]
\caption{Results over the GPT2 models family. The left number is the base model and the right number is with our pruning approach. Results are averaged across 5 different runs.}
\centering
\footnotesize
\resizebox{\textwidth}{!}{
\begin{tabular}{lcccccccc}
\multirow{2}{*}{\textbf{Task}} & \multicolumn{4}{c}{\textbf{GPT2-Small}} & \multicolumn{4}{c}{\textbf{GPT2-Medium}} \\
& \textbf{1-shot} & \textbf{2-shot} & \textbf{3-shot} & \textbf{4-shot} & \textbf{1-shot} & \textbf{2-shot} & \textbf{3-shot} & \textbf{4-shot} \\
\hline
T1 & .35$|$\textbf{.50} & .94$|$\textbf{.98} & .98$|$\textbf{1.0} & .99$|$\textbf{1.0} & .36$|$\textbf{.72} & .91$|$\textbf{.98} & .98$|$\textbf{1.0} & .99$|$\textbf{1.0} \\
T2 & .24$|$\textbf{.53} & .96$|$.96 & 1.0$|$1.0 & 1.0$|$1.0 & .61$|$\textbf{.78} & .99$|$\textbf{1.0} & .99$|$\textbf{1.0} & 1.0$|$1.0 \\
T3 & .42$|$.42 & .62$|$\textbf{.67} & .63$|$\textbf{.68} & .62$|$\textbf{.65} & .73$|$\textbf{.87} & .82$|$\textbf{.87} & .86$|$\textbf{.90} & .87$|$\textbf{.92} \\
T4 & .42$|$\textbf{.62} & .67$|$.67 & .68$|$.68 & .68$|$.68 &  \textbf{.66}$|$.62& \textbf{.78}$|$.75 & .82$|$.82 & \textbf{.83}$|$.80 \\
T5 & .57$|$\textbf{.61} & .53$|$\textbf{.59} & .52$|$\textbf{.57} &  .52$|$\textbf{.57}& .54$|$.54 & .51$|$\textbf{.53} & .52$|$\textbf{.54} & .54$|$\textbf{.56} \\
T6 & .30$|$.30 & .34$|$\textbf{.38} & .38$|$\textbf{.44} & .40$|$\textbf{.45} & .35$|$\textbf{.39} & .37$|$\textbf{.42} & .41$|$\textbf{.49} & .44$|$\textbf{.52} \\
T7 & .02$|$\textbf{.15} & .17$|$\textbf{.55} & .34$|$\textbf{.64} & .46$|$\textbf{.68} & .40$|$\textbf{.54} & .38$|$\textbf{.54} & .32$|$\textbf{.44} & .35$|$\textbf{.45} \\
T8 & .10$|$\textbf{.17} & .25$|$\textbf{.38} & .26$|$\textbf{.38} & .28$|$\textbf{.40} & .16$|$\textbf{.21} & .28$|$\textbf{.38} & .32$|$\textbf{.45} & .36$|$\textbf{54} \\
T9 & .10$|$\textbf{.30} & .29$|$\textbf{.49} & .35$|$\textbf{.52} & .43$|$\textbf{.57} & .23$|$\textbf{.30} & .47$|$\textbf{.55} & .55$|$\textbf{.64} & .63$|$\textbf{.72} \\
T10 & .27$|$\textbf{.30} & .35$|$\textbf{.39} & .36$|$\textbf{.40} & .37$|$\textbf{.42} & .34$|$\textbf{.38} & .45$|$\textbf{.48} & .54$|$.54 & .59$|$.59 \\
T11 & .06$|$\textbf{.19} & .10$|$\textbf{.29} & .11$|$\textbf{.27} & .11$|$\textbf{.25} & .49$|$\textbf{.52} & .62$|$\textbf{.65} & .67$|$.67 & .67$|$.67 \\
T12 & .19$|$\textbf{.26} & .33$|$\textbf{.44} & .38$|$\textbf{46} & .40$|$\textbf{.49} & .26$|$\textbf{.31} &  .33$|$\textbf{.42}& .36$|$\textbf{.47} &  .38$|$\textbf{48}\\
T13 & .18$|$.18 & .16$|$.16 & .18$|$.18 & .20$|$.20 & .17$|$.17 & .24$|$.24 & .26$|$.26 & .26$|$\textbf{.27} \\
T14 &.28$|$\textbf{.30}  & .38$|$\textbf{.40} & .40$|$\textbf{.43} & .42$|$\textbf{.44} & .31$|$.31 & .41$|$.41 & .45$|$\textbf{.47}  & .47$|$.47 \\
T15 & .11$|$.11 & .22$|$.22 & .24$|$.24 & .25$|$.25 & .19$|$\textbf{.22} & .30$|$.30 & .32$|$.32 & \textbf{.36}$|$.35 \\
T16 & .26$|$\textbf{.30} & .35$|$.35 & .33$|$.33 & .33$|$.33 & .41$|$\textbf{.45} & .47$|$.47 & \textbf{.47}$|$.45 & \textbf{.49}$|$.48 \\
T17 & .49$|$\textbf{.54} & .64$|$\textbf{.68} & .66$|$\textbf{.70} & .67$|$\textbf{.71} & .41$|$\textbf{.45} & .55$|$\textbf{.57} & .49$|$\textbf{.55} & .51$|$\textbf{.58} \\
T18 & .52$|$\textbf{.54} & .48$|$\textbf{.53} & .61$|$\textbf{.63} & .70$|$\textbf{.72} & .54$|$.54 & .55$|$.55 & .59$|$\textbf{.60} & .62$|$\textbf{.64} \\
\end{tabular}
}
\label{tab:transposed_tasks_gpt}
\end{table*}

\begin{table*}[h]
\caption{Results over the BLOOM models family. The left number is the base model and the right number is with our pruning approach. Results are averaged across 5 different runs.}
\centering
\footnotesize
\resizebox{\textwidth}{!}{
\begin{tabular}{lcccccccc}
\multirow{2}{*}{\textbf{Task}} & \multicolumn{4}{c}{\textbf{Bloom-560M}} & \multicolumn{4}{c}{\textbf{Bloom-1.1B}} \\
& \textbf{1-shot} & \textbf{2-shot} & \textbf{3-shot} & \textbf{4-shot} & \textbf{1-shot} & \textbf{2-shot} & \textbf{3-shot} & \textbf{4-shot} \\
\hline
T1 & .34$|$\textbf{.57} & .86$|$\textbf{.92} & .98$|$\textbf{1.0} & .99$|$\textbf{1.0} & .21$|$\textbf{.25} & .66$|$\textbf{.70} & .85$|$\textbf{.90} & .93$|$\textbf{.95} \\
T2 & .76$|$\textbf{.81} & .92$|$\textbf{.96} & .96$|$\textbf{.98} & .96$|$\textbf{.98} & .65$|$\textbf{.69} & .99$|$.99 & 1.0$|$1.0 & 1.0$|$1.0 \\
T3 & .37$|$\textbf{.46} & .71$|$\textbf{.77} & .83$|$\textbf{.87} & .87$|$\textbf{.90} & .43$|$\textbf{.48} & .83$|$\textbf{.85} & .92$|$\textbf{.93} & .93$|$\textbf{.95} \\
T4 & .43$|$\textbf{.54} & .62$|$\textbf{.71} & .68$|$\textbf{.78} & .72$|$\textbf{.83} & .52$|$\textbf{.56} & .75$|$.75 & .77$|$.77 & \textbf{.83}$|$.80 \\
T5 & .51$|$\textbf{.54} & .51$|$.51 & .56$|$.56 & .57$|$.57 & .58$|$\textbf{.60} & .62$|$.62 & .60$|$\textbf{.63} & .60$|$.60 \\
T6 & .22$|$\textbf{.27} & .33$|$\textbf{.36} & .38$|$\textbf{.42} & .40$|$\textbf{.42} & .24$|$.24 & .28$|$.28 & .33$|$.33 & .34$|$.34 \\
T7 & .15$|$\textbf{.19} & .24$|$\textbf{.28} & .28$|$.28 & \textbf{.33}$|$.32 & .21$|$\textbf{.32} & .41$|$\textbf{.60} & .44$|$\textbf{.65} & .51$|$\textbf{.73} \\
T8 & .36$|$\textbf{.47} & .44$|$\textbf{.58} & .49$|$\textbf{.62} & .58$|$\textbf{.67} & .48$|$\textbf{.57} & .65$|$\textbf{.72} & .74$|$\textbf{.76} & .81$|$\textbf{.83} \\
T9 & .35$|$\textbf{.42} & .45$|$\textbf{.56} & .56$|$\textbf{.64} & .57$|$\textbf{.66} & .53$|$\textbf{.58} & .73$|$\textbf{.76} & .81$|$\textbf{.85} & .84$|$.84 \\
T10 & .30$|$\textbf{.36} & .42$|$\textbf{.49} & .41$|$\textbf{.48} & .45$|$\textbf{.52} & .40$|$\textbf{.44} & .52$|$\textbf{.55} & .56$|$\textbf{.60} & .62$|$\textbf{.66} \\
T11 & .42$|$\textbf{.47} & .56$|$\textbf{.61} & .60$|$\textbf{.64} & .65$|$.65 & .57$|$\textbf{.61} & .68$|$.68 & .71$|$.71 & .73$|$\textbf{.77} \\
T12 & .39$|$\textbf{.46} & .38$|$\textbf{.50} & .45$|$\textbf{.54} & .42$|$\textbf{.51} & .40$|$\textbf{.47} & .48$|$\textbf{.56} & .62$|$\textbf{.68} & .68$|$\textbf{74} \\
T13 & .16$|$\textbf{.18} & .22$|$.22 & .21$|$.21 & .23$|$\textbf{.25} & .29$|$\textbf{.31} & .40$|$.40 & .41$|$.41 & .42$|$.42 \\
T14 & .26$|$\textbf{.28} & .39$|$\textbf{.41} & .45$|$\textbf{.48} & .41$|$\textbf{.43} & .36$|$.36 & .48$|$.48 & .44$|$\textbf{.46} & .47$|$.47 \\
T15 & .18$|$\textbf{.22} & .26$|$\textbf{.29} & .29$|$\textbf{.33} & .29$|$\textbf{.31} & .28$|$\textbf{.32} & .37$|$\textbf{.42} & .40$|$\textbf{.45} & .42$|$.42 \\
T16 & .24$|$\textbf{.26} & .36$|$\textbf{.39} & .32$|$\textbf{.34} & .30$|$.30 & .52$|$.52 & .58$|$.58 & .57$|$\textbf{.61} & .57$|$\textbf{.62} \\
T17 & .49$|$\textbf{.52} & \textbf{.45}$|$.44 & \textbf{.57}$|$.55 & \textbf{.63}$|$.61 & .57$|$\textbf{.60} & .63$|$\textbf{.67} & .66$|$.66 & .70$|$.70 \\
T18 & .59$|$.59 & .55$|$.55 & \textbf{.64}$|$.62 & \textbf{.66}$|$.63 & .49$|$\textbf{.51} & .53$|$\textbf{.55} & .64$|$.64 & .65$|$\textbf{.67} \\
\end{tabular}
}
\label{tab:transposed_tasks_bloom}
\end{table*}

\begin{table*}[h]
\caption{Results over the OPT models family, the left number is the base model and the right number is with our pruning approach, results are averaged across 5 different runs}
\label{tab:tasks_opt_long}
\footnotesize
\centering
\resizebox{\textwidth}{!}{
\begin{tabular}{lcccccccc}
& \multicolumn{4}{c}{\textbf{OPT-125M}} & \multicolumn{4}{c}{\textbf{OPT-350M}} \\
\textbf{Task} & \textbf{1 Shot} & \textbf{2 Shot} & \textbf{3 Shot} & \textbf{4 Shot} & \textbf{1 Shot} & \textbf{2 Shot} & \textbf{3 Shot} & \textbf{4 Shot} \\
\hline
T1 & .01$|$\textbf{.18} & .37$|$\textbf{.59} & .21$|$\textbf{.35} & .21$|$\textbf{.31} & .02$|$\textbf{.15} & .23$|$\textbf{.55} & .52$|$\textbf{.81} & .72$|$\textbf{.90} \\
T2 & .02$|$\textbf{.17} & .27$|$\textbf{.49} & .25$|$\textbf{.28} & .21$|$\textbf{.24} & .24$|$\textbf{.51} & .72$|$\textbf{.87} & .85$|$\textbf{.94} & .92$|$\textbf{.97} \\
T3 & .19$|$\textbf{.34} & .30$|$\textbf{.38} & .32$|$\textbf{.38} & .32$|$\textbf{.39} & .21$|$\textbf{.29} & .55$|$\textbf{.64} & .62$|$\textbf{.64} & \textbf{.61}$|$.60 \\
T4 & .18$|$\textbf{.37} & .30$|$\textbf{.40} & .30$|$\textbf{.40} & .29$|$\textbf{.39} & .55$|$.55 & \textbf{.36}$|$.34 & \textbf{.33}$|$.30 & .27$|$.27 \\
T5 &.32$|$\textbf{.39}& .33$|$.33 & \textbf{.30}$|$.29 & .30$|$.30 & .49$|$\textbf{.51} & .46$|$.46 & .46$|$.46 & .47$|$\textbf{.49} \\
T6 & .25$|$\textbf{.35} & .31$|$\textbf{.35} & .30$|$\textbf{.34} & .33$|$\textbf{.35} & \textbf{.30}$|$.29 & \textbf{.41}$|$.39 & \textbf{.40}$|$.38 & \textbf{.44}$|$.41 \\
T7 &.01$|$\textbf{.07}& .03$|$\textbf{.12} & .03$|$\textbf{.09} & .04$|$\textbf{.08} & .03$|$\textbf{.05} & .15$|$\textbf{.30} & .15$|$\textbf{.28} & .19$|$\textbf{.31} \\
T8 & .01$|$.01 & .03$|$\textbf{.05} & .04$|$\textbf{.07} & .04$|$\textbf{.06} & .05$|$\textbf{.07} & .12$|$\textbf{.15} & .17$|$\textbf{.21} & .19$|$\textbf{.30} \\
T9 & .01$|$\textbf{.06} & .11$|$\textbf{.16} & .14$|$\textbf{.18} & .15$|$\textbf{.21} & .07$|$.07 & .19$|$\textbf{.22} & .23$|$\textbf{.25} & .23$|$\textbf{.28} \\
T10 & .10$|$\textbf{.19} & .22$|$\textbf{.25} & .28$|$\textbf{.30} & .29$|$\textbf{.32} & .28$|$\textbf{.33} & .35$|$\textbf{.37} & .39$|$\textbf{.41} & .38$|$\textbf{.40} \\
T11 & .05$|$\textbf{.10} & .11$|$\textbf{.15} & .15$|$\textbf{.17} & .16$|$\textbf{.18} & .20$|$\textbf{.29} & .32$|$\textbf{.38} & .33$|$\textbf{.37} & .29$|$\textbf{.33} \\
T12 & .02$|$\textbf{.09} & .12$|$\textbf{.16} & .16$|$.16 & .16$|$.16 & .16$|$\textbf{.19} & .24$|$\textbf{.26} & .27$|$\textbf{.29} & .26$|$\textbf{.28} \\
T13 & .08$|$\textbf{.12} & .07$|$\textbf{.12} & .10$|$\textbf{.15} & .11\textbf{.16} & .14$|$\textbf{.16} & .24$|$\textbf{.27} & .32$|$\textbf{.34} & .32$|$.32 \\
T14 & .09$|$\textbf{.25} & .26$|$\textbf{.34} & .31$|$\textbf{.33} & .31$|$\textbf{.34} & .33$|$\textbf{.37} & .48$|$\textbf{.50} & .42$|$\textbf{.44} &  .51$|$\textbf{.54}\\
T15 & .04$|$\textbf{.09} & .13$|$\textbf{.16} & .15$|$\textbf{.18} & .15$|$\textbf{.17} & .11$|$\textbf{.13} & .20$|$.20 & .24$|$\textbf{.25} & .25$|$\textbf{.26} \\
T16 & .19$|$\textbf{.24} & .26$|$\textbf{.30} & .25$|$\textbf{.28} & .24$|$\textbf{.26} & \textbf{.29}$|$.28 & \textbf{.42}$|$.41 & \textbf{.45}$|$.44 & \textbf{.46}$|$.45 \\
T17 & .54$|$.54 & .54$|$\textbf{.57} & .53$|$\textbf{.55} & .62$|$.62 & \textbf{.51}$|$.50 & .64$|$.64 & .67$|$.67 & .72$|$.72 \\
T18 & .56$|$.56 & \textbf{.65}$|$.63 & .65$|$.65 & .69$|$.69 & .51$|$\textbf{.53} & .47$|$\textbf{.50}  & .56$|$\textbf{.58} & .60$|$\textbf{.67}
\\ \\
& \multicolumn{4}{c}{\textbf{OPT-1.3B}} & \multicolumn{4}{c}{\textbf{OPT-2.7B}} \\
\textbf{Task} & \textbf{1 Shot} & \textbf{2 Shot} & \textbf{3 Shot} & \textbf{4 Shot} & \textbf{1 Shot} & \textbf{2 Shot} & \textbf{3 Shot} & \textbf{4 Shot} \\
\hline
T1 & .01$|$\textbf{.13} & .24$|$\textbf{.83} & .76$|$\textbf{.98} & .94$|$\textbf{1.0} & .09$|$\textbf{.13} & .89$|$\textbf{.98} & .99$|$\textbf{1.0} & 1.0$|$1.0 \\
T2 & .02$|$\textbf{.14} & .73$|$\textbf{.82} & .92$|$\textbf{.97} & .97$|$\textbf{.99} & .04$|$\textbf{.12} & .72$|$\textbf{.93} & .86$|$\textbf{.98} & .91$|$\textbf{1.0} \\
T3 & .78$|$\textbf{.93} & .91$|$.91 & .92$|$.92 & \textbf{.89}$|$.88 & .84$|$\textbf{.90} & .93$|$\textbf{.95} & .97$|$\textbf{1.0} & .98$|$\textbf{1.0} \\
T4 &.37$|$\textbf{.56}  & .62$|$\textbf{.74} & .59$|$\textbf{.71} & .68$|$\textbf{.74} & .47$|$\textbf{.57} & .64$|$\textbf{.76} & .77$|$\textbf{.80} & .74$|$\textbf{.76} \\
T5 & .67$|$\textbf{.70} & .61$|$\textbf{.64} & .65$|$\textbf{.67} & .62$|$\textbf{.68} & .56$|$.56 & .49$|$.49 & .54$|$.54 & .55$|$.55 \\
T6 & .29$|$\textbf{.31} & .43$|$\textbf{.45} & .51$|$\textbf{.53} & .54$|$\textbf{.59} & .35$|$\textbf{.37} & .47$|$.47 & .51$|$\textbf{.52} & .54$|$.54 \\
T7 & .28$|$\textbf{.34} & .52$|$\textbf{.65} & .53$|$\textbf{.60} & .56$|$.56 & .14$|$\textbf{.23} & .44$|$\textbf{.50} & .50$|$\textbf{.58} & .44$|$\textbf{.49} \\
T8 & .11$|$\textbf{.16} & .29$|$\textbf{.45} & .42$|$\textbf{.60} & .57$|$\textbf{.70} & .20$|$\textbf{.32} & .49$|$\textbf{.65} & .61$|$\textbf{.73} & .68$|$\textbf{.75} \\
T9 & .12$|$\textbf{.14} & .27$|$\textbf{.45} & .49$|$\textbf{.63} & .59$|$\textbf{.71} & .28$|$\textbf{.39} & .68$|$\textbf{.75} & .80$|$\textbf{.83} & .82$|$\textbf{.85} \\
T10 & .34$|$\textbf{.36} & .36$|$\textbf{.41} & .42$|$\textbf{.45} & .45$|$\textbf{.49} & .35$|$\textbf{.40} & .46$|$\textbf{.50} & .52$|$\textbf{.56} & .59$|$\textbf{.62} \\
T11 & .60$|$\textbf{.62} & .70$|$\textbf{.71} & .74$|$.74 & .75$|$.75 & .64$|$.64 & .75$|$\textbf{.77} & .76$|$\textbf{.78} & .78$|$.78 \\
T12 & .27$|$\textbf{.30} & .30$|$\textbf{.37} & .38$|$\textbf{.43} & .41$|$\textbf{.49} & .28$|$\textbf{.33} & .40$|$\textbf{.47} & .42$|$\textbf{.44} & .47$|$\textbf{.52} \\
T13 & .32$|$\textbf{.38} & .38$|$\textbf{.42} & .46$|$\textbf{.48} & .45$|$\textbf{.49} & .42$|$\textbf{.46} & .50$|$\textbf{.52} & .55$|$\textbf{.57} & .61$|$.61 \\
T14 & .41$|$.41 & .48$|$.48 & .51$|$.51 & .51$|$.51 & .43$|$.43 & .53$|$\textbf{.55} & .49$|$\textbf{.50} & .52$|$.52 \\
T15 & .24$|$\textbf{.26} & .38$|$\textbf{.40} & .42$|$\textbf{.44} & .44$|$\textbf{.47} & .25$|$\textbf{.30} & .40$|$\textbf{.45} & .50$|$\textbf{.54} & .53$|$\textbf{.57} \\
T16 & .56$|$.56 & .65$|$.65 & .65$|$.65 & .68$|$.68 & .70$|$.70 & .74$|$.74 & .72$|$.72 & .74$|$.74 \\
T17 & .55$|$\textbf{.60} & .67$|$\textbf{.68} & .73$|$\textbf{.73} & .77$|$.77 & .54$|$\textbf{.59} & .62$|$\textbf{.64} & .72$|$.72 & .72$|$.72 \\
T18 & .55$|$.55 & .61$|$.61 & .65$|$.65 & .71$|$.71 & .48$|$\textbf{.50} & .59$|$.59 & .69$|$.69 & .73$|$.73 \\
\hline
\end{tabular}}
\end{table*}
\newpage

\begin{table*}[!t]
\caption{Results over the Mamba models family. The left number is the base model and the right number is with our pruning approach. Results are averaged across 5 different runs.}
\centering
\footnotesize
\resizebox{\textwidth}{!}{
\begin{tabular}{lcccccccc}
\multirow{2}{*}{\textbf{Task}} & \multicolumn{4}{c}{\textbf{Mamba-130M}} & \multicolumn{4}{c}{\textbf{Mamba-370M}} \\
& \textbf{1-shot} & \textbf{2-shot} & \textbf{3-shot} & \textbf{4-shot} & \textbf{1-shot} & \textbf{2-shot} & \textbf{3-shot} & \textbf{4-shot} \\
\hline
T1 & .77$|$\textbf{.97} & 1.0$|$1.0 & 1.0$|$1.0 & 1.0$|$1.0 & .95$|$\textbf{.99} & 1.0$|$1.0 & 1.0$|$1.0 &  1.0$|$1.0 \\
T2 & .09$|$\textbf{.73} & .95$|$\textbf{1.0} & 1.0$|$1.0 & 1.0$|$1.0 & .55$|$\textbf{.83} & .88$|$\textbf{.92} & 1.0$|$1.0 & 1.0$|$1.0 \\
T3 & .39$|$\textbf{.72} & .82$|$\textbf{.95} & .89$|$\textbf{1.0} & .92$|$\textbf{1.0} & .75$|$\textbf{.79} & .98$|$\textbf{.99} & .99$|$\textbf{1.0} & 1.0$|$1.0 \\
T4 & .73$|$.73 & .86$|$\textbf{.88} & .92$|$.92 & .93$|$.93 & .49$|$\textbf{.53} & .80$|$.80 & .86$|$.86 & 0.91$|$\textbf{.93} \\
T5 & .47$|$\textbf{.50} & .49$|$\textbf{.53} & .53$|$\textbf{.55} & .53$|$\textbf{.56} & .53$|$\textbf{.56} & .51$|$\textbf{.53} & .62$|$\textbf{.64} & .60$|$\textbf{.62} \\
T6 & .32$|$\textbf{.35} & .46$|$\textbf{.48} & .47$|$\textbf{.50} & .47$|$\textbf{.52} & .34$|$\textbf{.38} & .41$|$\textbf{.43} & .51$|$\textbf{.53} & .54$|$\textbf{.55} \\
T7 & .14$|$.14 & .70$|$.70 & .73$|$.73 & .81$|$\textbf{.83} & .60$|$\textbf{.69} & .87$|$\textbf{.89} & .88$|$\textbf{.90} & .85$|$.85 \\
T8 & .08$|$\textbf{.20} & .36$|$\textbf{.48} & .42$|$\textbf{.54} & .51$|$\textbf{.57} & .38$|$\textbf{45} & .56$|$\textbf{.59} & .63$|$.63 & .70$|$.70  \\
T9 &  .11$|$\textbf{.25}& .36$|$\textbf{.50} & .50$|$\textbf{.55} & .60$|$\textbf{.65} & .55$|$\textbf{.59} & .74$|$\textbf{.77} & .82$|$\textbf{.85} & .86$|$\textbf{.88} \\
T10 & .25$|$.25 & .30$|$.30 & .39$|$\textbf{.42} & .48$|$\textbf{.50} & .34$|$\textbf{.41} & .46$|$\textbf{.52} & .56$|$\textbf{.58} & .68$|$\textbf{.70} \\
T11 & .20$|$.20 & .29$|$.29 & .35$|$.35 & .30$|$.30 & .45$|$\textbf{.51} & .61$|$\textbf{.64} & .70$|$\textbf{.73} & .66$|$\textbf{.69} \\
T12 & .19$|$\textbf{.22} & .36$|$\textbf{.43} & .47$|$\textbf{.52} & .59$|$\textbf{.62} & .32$|$\textbf{.40} & .35$|$\textbf{.42} & .45$|$\textbf{.49} & .49$|$\textbf{.52} \\
T13 & .16$|$\textbf{.20} & .26$|$.26 & .28$|$\textbf{.30} & .36$|$.36 & .37$|$\textbf{.41} & .52$|$\textbf{.55} & .54$|$\textbf{.57} & .53$|$\textbf{.55} \\
T14 & .28$|$\textbf{.33} & .33$|$\textbf{.36}& .36$|$\textbf{.40} & .40$|$\textbf{.42} & .35$|$.35 & .41$|$.41 & .46$|$\textbf{.48} & .46$|$\textbf{.48} \\
T15 & .08$|$\textbf{.10} & .14$|$.14 & .19$|$.19 & .20$|$.20 & .31$|$\textbf{.36} & .45$|$\textbf{.49} & .54$|$.54 & .56$|$.56 \\
T16 & .39$|$\textbf{.44} & .50$|$.50 & .60$|$.60 & .63$|$.63 & .62$|$\textbf{.66} & .68$|$\textbf{.72} & .70$|$\textbf{.73} & .72$|$\textbf{.75} \\
T17 & .52$|$\textbf{.55} & .65$|$\textbf{.67} & .61$|$\textbf{.65} & .64$|$\textbf{.70} & .51$|$\textbf{.57} & .58$|$\textbf{.61} & .58$|$\textbf{.61} & .58$|$\textbf{.60} \\
T18 & .48$|$\textbf{.50} & .48$|$\textbf{.50} & .56$|$.56 & .51$|$\textbf{.53} & .53$|$\textbf{.56} & .51$|$\textbf{.55} & .59$|$.59 & .60$|$.60 \\

\end{tabular}
}
\label{tab:transposed_tasks_mamba}
\end{table*}
\newpage
\section{Task Vectors}
We provide additional results for the task-vectors experiment, we include two additional models (GPT2-Small and Bloom1.1B) over the Algorithmic Next Letter task and Linguistics Antonyms.
\label{appendix:taskvectors}
.
\begin{figure*}[ht]
\centering
\includegraphics[width=1\textwidth,height=0.37\textwidth]{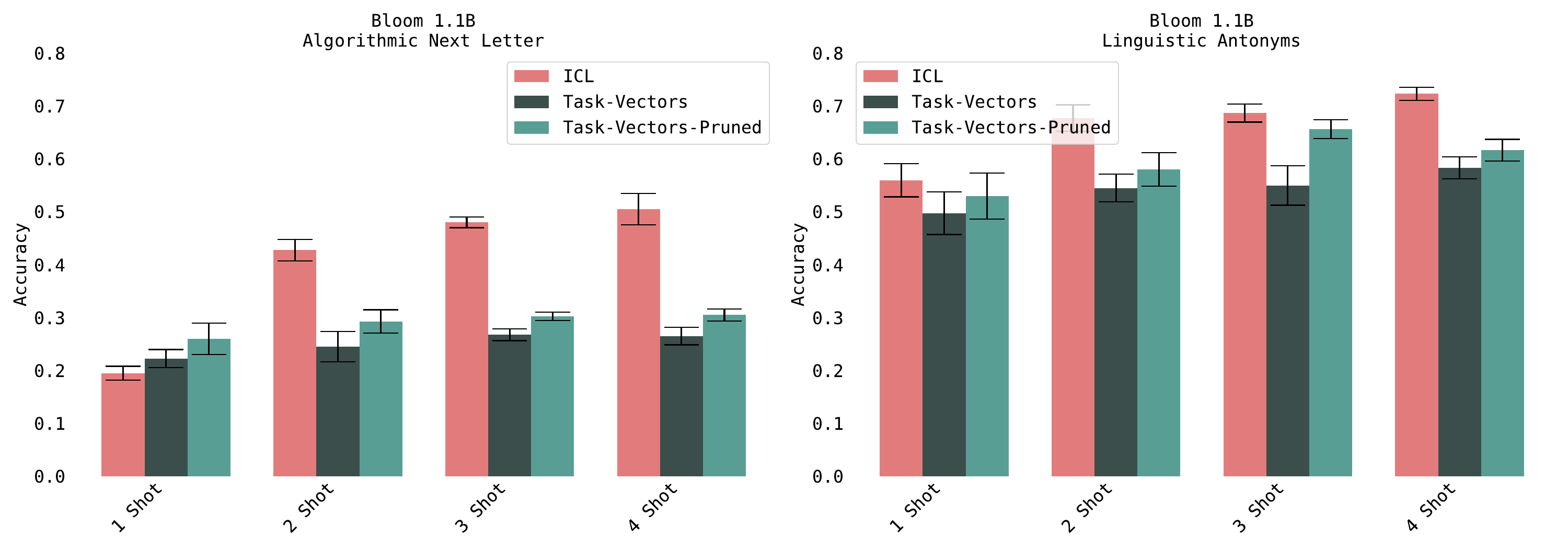}
\\
\includegraphics[width=1\textwidth,height=0.37\textwidth]{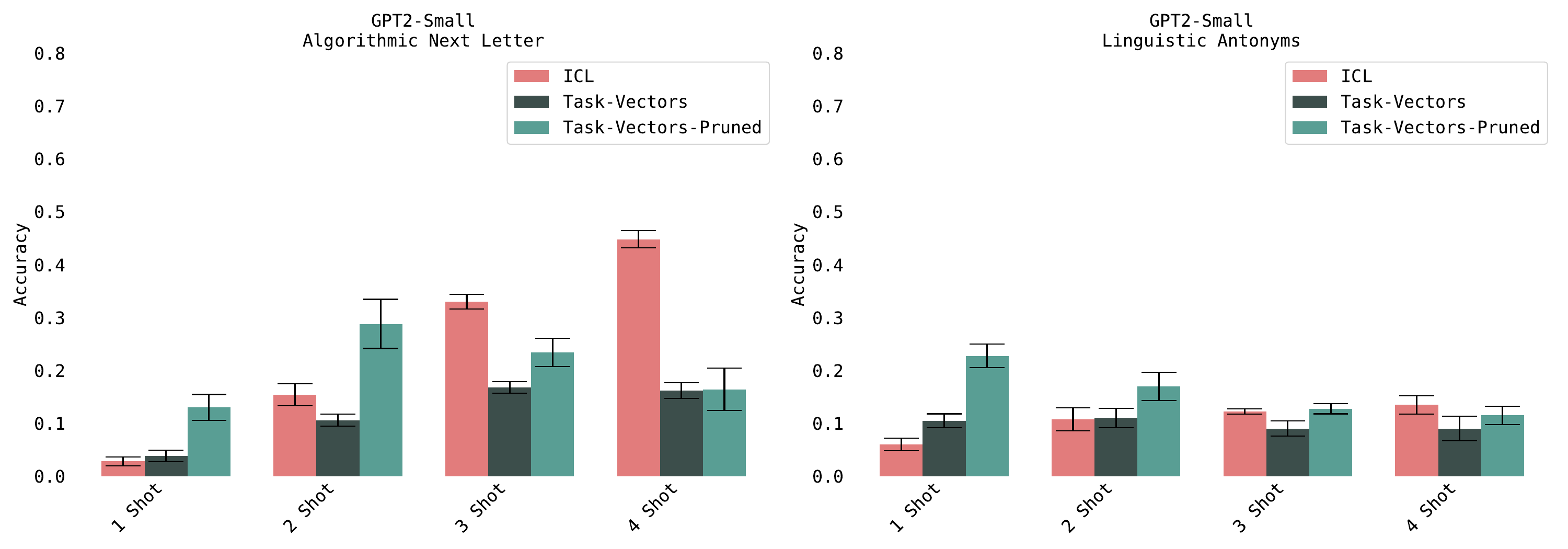}
\caption{Additional quantitative results for the task-vectors experiment }
\label{fig:task_vectors_2}
\end{figure*}
\end{document}